\documentclass{article} % For LaTeX2e
\usepackage{iclr2022_conference,times}

\usepackage{amsmath,amsfonts,bm}

\def\eqref#1{equation~\ref{#1}}
\def\1{\bm{1}}

\DeclareMathAlphabet{\mathsfit}{\encodingdefault}{\sfdefault}{m}{sl}
\SetMathAlphabet{\mathsfit}{bold}{\encodingdefault}{\sfdefault}{bx}{n}

\usepackage{hyperref}
\usepackage{url}
\usepackage{algorithm}  
\usepackage{algorithmicx,algpseudocode}
\usepackage{stfloats}
\usepackage{xspace}
\usepackage{wrapfig}
\usepackage{lipsum}
\usepackage{setspace}
\usepackage{enumitem}
\usepackage{microtype}
\usepackage{graphicx}
\usepackage{subfigure}
\usepackage{booktabs} % for professional tables
\usepackage{amsfonts}
\usepackage{multirow}
\usepackage{stfloats}
\usepackage[misc]{ifsym}

\definecolor{MyDarkBlue}{rgb}{0,0.08,1}
\definecolor{MyDarkGreen}{rgb}{0.02,0.6,0.02}
\definecolor{MyDarkRed}{rgb}{0.8,0.02,0.02}
\definecolor{MyDarkOrange}{rgb}{0.40,0.2,0.02}
\definecolor{MyPurple}{RGB}{111,0,255}
\definecolor{MyRed}{rgb}{1.0,0.0,0.0}
\definecolor{MyGold}{rgb}{0.75,0.6,0.12}
\definecolor{MyDarkgray}{rgb}{0.66, 0.66, 0.66}

\newcommand{\myparagraph}[1]{\vspace{-5pt}\paragraph{#1}}

\title{ModeRNN: Harnessing Spatiotemporal Mode\\ Collapse in Unsupervised Predictive Learning}

\iclrfinalcopy

\author{
	Zhiyu Yao\textsuperscript{1}\thanks{Equal contribution}, Yunbo Wang\textsuperscript{2}\footnotemark[1],  Haixu Wu\textsuperscript{1}, Jianmin Wang\textsuperscript{1}, Mingsheng Long \textsuperscript{1} (\Letter) \\
	\textsuperscript{1}School of Software, BNRist, Tsinghua University, China \\	\textsuperscript{2}MoE Key Lab of Artificial Intelligence, AI Institute, Shanghai
	Jiao Tong University, China \\
	\textsuperscript{1}{\tt\small \{yaozy19,whx20\}@mails.tsinghua.edu.cn, \{jimwang,mingsheng\}@tsinghua.edu.cn},  \\
		\textsuperscript{2}{\tt\small yunbow@sjtu.edu.cn},
	 
}

\begin{document}

\maketitle

\begin{abstract}
	
	Learning predictive models for unlabeled spatiotemporal data is challenging in part because visual dynamics can be highly entangled in real scenes, making existing approaches prone to overfit partial modes of physical processes while neglecting to reason about others.
	We name this phenomenon \textit{spatiotemporal mode collapse} and explore it for the first time in predictive learning.
	The key is to provide the model with a strong inductive bias to discover the compositional structures of latent modes.
	To this end, we propose ModeRNN, which introduces a novel method to learn structured hidden representations between recurrent states.
	The core idea of this framework is to first extract various components of visual dynamics using a set of \textit{spatiotemporal slots} with independent parameters.
	Considering that multiple space-time patterns may co-exist in a sequence, we leverage learnable importance weights to adaptively aggregate slot features into a unified hidden representation, which is then used to update the recurrent states.
	Across the entire dataset, different modes result in different responses on the mixtures of slots, which enhances the ability of ModeRNN to build structured representations and thus prevents the so-called mode collapse. 
	Unlike existing models, ModeRNN is shown to prevent spatiotemporal mode collapse and further benefit from learning mixed visual dynamics. 
	
\end{abstract}

\section{Introduction}

\begin{figure}[t]
	\centering
	\subfigure[Learning RNNs under highly mixed modes of visual dynamics (\textit{e.g.}, collision-free vs. collision) may lead to \textit{spatiotemporal mode collapse}, \textit{i.e.}, the predictions of some modes are severely affected by the presence of others.]{
		\includegraphics[width=\textwidth]{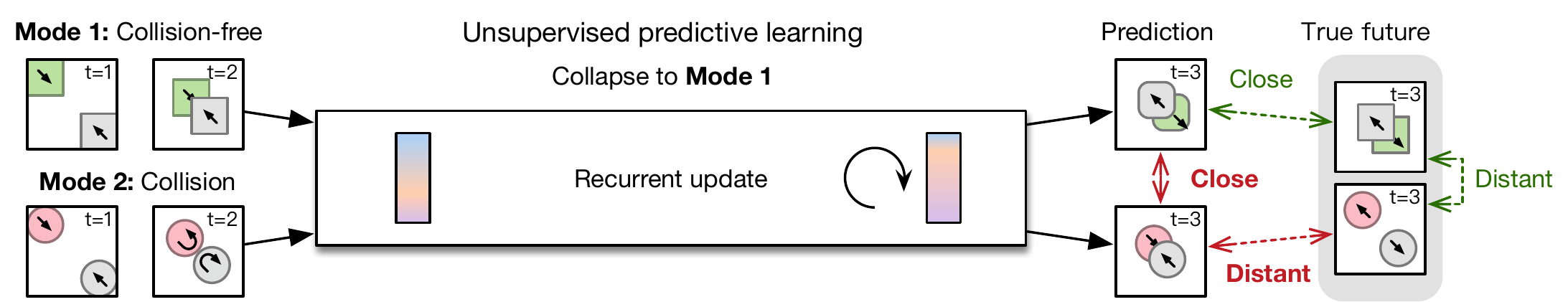}
		\label{fig:intro_rnn}
	}
	\subfigure[ModeRNN tackles mode collapse by learning the compositional structures of visual dynamics through a set of \textit{spatiotemporal slots}. The three modules in each recurrent unit of ModeRNN will be discussed later.]{
		\includegraphics[width=\textwidth]{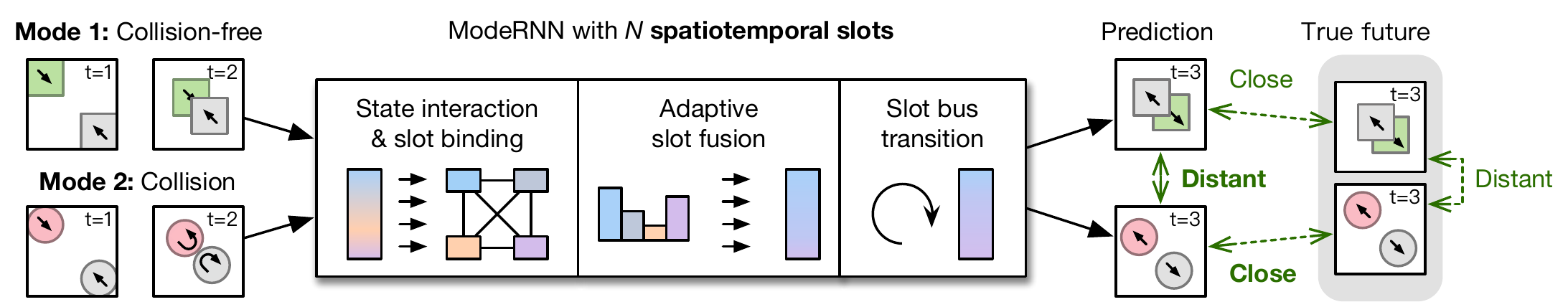}
		\label{fig:intro_modernn}
	}
	\vskip -0.1in
	\caption{Illustration of \textit{spatiotemporal mode collapse} and our approach.}
	\label{fig:intro_}
\end{figure}

Predictive learning is an unsupervised learning method that has shown the ability to discover latent structures of unlabeled spatiotemporal data. 
However, in practice, the spatiotemporal data often contains a variety of visual dynamics, which are mainly reflected in the richness of spatial correlations and movement trends, as well as the diversity of interactions between multiple objects (see Figure \ref{fig:intro_}).
It remains a challenge for existing predictive models to fully capture these \textit{spatiotemporal modes} in a completely unsupervised manner using regular forward modeling mechanisms, such as recurrent update \citep{shi2015convolutional}, autoregression \citep{Kalchbrenner2017Video}, and 3D convolutions \citep{wang2019eidetic}, while ignoring the inherent differences in dynamics among data samples.
For clarity, in the following discussion, \textit{spatiotemporal modes} are considered to have the following properties: 
\begin{enumerate}
	\item Multiple spatiotemporal modes naturally exist in the dataset and are unknown before model training. They can be viewed as prototypes of space-time representations. For simplicity, each sequence corresponds to a single spatiotemporal mode.
	\item Multiple modes share a set of hidden representation subspaces (referred to as ``\textit{spatiotemporal slots}'') and have different compositional structures over the spatiotemporal slots.
\end{enumerate}

We observe that if the predictive model experiences multiple complex modes in spacetime during training, the learning processes of different modes may affect each other, resulting in ambiguous prediction results. 
Figure \ref{fig:intro_rnn} provides an example that the predictive model responds effectively only to certain visual dynamics, from which it learns biased prior knowledge (\textit{e.g.}, collision-free), and generates future frames arbitrarily based on these priors while neglecting to reason about other possibilities of latent modes.
We name this phenomenon \textit{spatiotemporal mode collapse}.

Unlike the widely concerned \textit{mode collapse} problem in generative adversarial networks, the above issue is previously considered to be less likely to occur in the generation of future frames conditioned on sequential observations, because the training process is often constrained by the image reconstruction loss.
If, for some reason, the predictive model decided to concentrate on a single mode, fitting this mode with a large number of model parameters, then the training examples of other modes would have poor recovery.
However, due to the limited number of parameters, it is possible when the model cannot effectively infer the latent structures of various spatiotemporal modes that are highly mixed across the dataset;
As a result, its responses to different modes in the feature space tend to lose diversity and collapse to certain neighborhoods that may be close to the features of some modes (as shown in Figure \ref{fig:intro_rnn}) or may be the average of feature ranges of multiple modes.

We explore this issue for the first time in predictive learning.
The core idea is to provide a strong inductive bias for the predictive model to discover the compositional structures of latent modes.
To this end, we propose ModeRNN, a new modular RNN architecture that learns structured hidden representations simply from unlabeled spatiotemporal data. 
Specifically, ModeRNN is built upon a set of \textit{spatiotemporal slots}\footnote{The concept of ``\textit{slot}'' was initially introduced by \citet{locatello2020object} to denote the object-centric representation in static scene understanding. We borrow this term here to refer to the subspaces of spatiotemporal features in dynamic visual scenes.} that respond to different components of mixed visual dynamics.
As shown in Figure \ref{fig:intro_modernn}, they are processed in three stages in each ModeRNN unit, following a decoupling-aggregation framework based on slot features, which is completely different from the existing predictive networks with modular architectures \citep{goyal2019recurrent,xu2019unsupervised}.
The first stage is recurrent state interaction and slot binding. We use the multi-head attention mechanism to enable the memory state to interact with the input state and previous hidden state of RNNs.
We name the memory state ``\textit{slot bus}'', because for each sequence, it is initialized from a multi-variate Gaussian distribution with learnable parameters, and thereafter refined using the slot features at each time step.
By using the slot bus as the queries, multi-head attention can naturally decouple modular components from hidden representations and bind them to particular spatiotemporal slots.
Features in each slot are then independently modeled using per-slot convolutional parameters.
The second stage in each ModeRNN unit is slot fusion, motivated by the assumptions that, first, there can be multiple patterns of visual dynamics in a single sequence; Second, different spatiotemporal modes have different compositional structures over the spatiotemporal slots.
Therefore, we assign slot features with learnable importance weights and aggregate them into a unified hidden representation, which is then used in the third stage to update the slot bus and generate the output state of the ModeRNN unit.

We show the existence of \textit{spatiotemporal mode collapse} on three datasets that are commonly used by previous literature. 
The mixed Moving MNIST dataset contains diverse space-time deformations due to different numbers of flying digits. The KTH dataset naturally contains multiple modes of visual structures in six action categories. The radar echo dataset for precipitation forecasting contains various modes that are reflected in a seasonal climate.
Through a series of quantitative and visualization results, we demonstrate the effectiveness of ModeRNN in learning from highly mixed visual dynamics.

\vspace{-5pt}
\section{Related Work}
% why spacetimee RNN
%\vspace{-5pt}
% \paragraph{Filter bank in signal processing.}
% \textit{Filter bank} \cite{10.5555/248702} is a classic method for intricate signal processing and has been widely used in audio and time series applications \cite{johnston1998audio,flandrin2004empirical}. It employs a series of empirical selected filters that separate the input signal into multiple components, each carrying a single frequency sub-band of the signal. Inspired by this classic work, we design a learnable mixture of filter structure for learning mixing dynamics from diverse spacetime sequences.
%\vspace{-5pt}
%\subsection{Spacetime RNNs}
%\vspace{-5pt}
\myparagraph{RNN-based predictive models.}
%\vspace{-5pt}
Many deep learning models based on RNNs have been proposed for spatiotemporal prediction \citep{Ranzato2014Video,srivastava2015unsupervised,shi2015convolutional,Oh2015Action,de2016dynamic, villegas2017learning,oliu2018folded,WangZZLWY19,wang2019eidetic,castrejon2019improved,pmlr-v119-yao20a,guen2020disentangling, yu2019efficient}. 
\citet{shi2015convolutional} integrated 2D convolutions into the recurrent state transitions of standard LSTM and proposed the convolutional LSTM network, which can model the spatial correlations and temporal dynamics in a unified recurrent unit.
\citet{wang2017predrnn} extended convolutional LSTMs with pairwise memory cells to capture both long-term dependencies and short-term variations to improve the prediction quality.
\citet{su2020convolutional} introduced a high-order convolutional tensor-train decomposition model named Conv-TT-LSTM to combine convolutional features across time for long-term prediction. 
In addition to the deterministic models, recent literature \citep{Mathieu2015Deep,vondrick2016generating,tulyakov2018mocogan, xu2018video, wang2018video,denton2018stochastic,kwon2019predicting, bhagat2020disentangling} also proposed probabilistic models that explicitly consider the uncertainty in predicting future sequences. For example, \citet{denton2018stochastic} introduced a stochastic video generation framework based on the conditional VAE architecture.
Different from above models, our approach is more focused on finding solutions to spatiotemporal mode collapse.

\myparagraph{Unsupervised predictive learning for spatiotemporal disentanglement.}
Previous work has focused on learning to disentangle the spatial and temporal features from visual dynamics \citep{denton2017unsupervised,guen2020disentangling,hsieh2018learning}.
These methods assume that the spatial information is temporally invariant, and factorize spatiotemporal data into two feature subspaces with strong semantic priors. 
Another line of work is to learn predictive models for unsupervised scene decomposition such as \citep{xu2019unsupervised,hsieh2018learning}. 
Unlike the above models, our approach uses a set of modular architectures in the recurrent unit to represent the mixed spatiotemporal dynamics.
The most relevant work to our method is the \textit{Recurrent Independent Mechanism} (RIM) \citep{goyal2019recurrent}, which consists of largely independent recurrent modules that are sparsely activated and interact via soft attention. 
ModeRNN is different from RIM in three aspects. First, it mainly tackles the problem of spatiotemporal mode collapse in a real-world environment.
Second, ModeRNN learns modular features by incorporating multi-head attention in the recurrent unit, and performs state transitions on compositional features with learnable importance weights.
Third, the modular structures (\textit{i.e.}, slots) in ModeRNN are frequently activated responding to the mixed visual dynamics.

\section{ModeRNN}

We propose ModeRNN to reduce spatiotemporal mode collapse in unsupervised predictive learning.
The key idea is that different latent modes in the same data domain should share a set of hidden representation subspaces which we call \textit{spatiotemporal slots}, and can be represented by different compositional structures based on the slot features. 
In this section, we first discuss the basic network components in ModeRNN and then describe the detailed architectures in each recurrent unit.

\subsection{Spatiotemporal Slots \& Slot Bus}

As described earlier, the term \textit{spatiotemporal slot} is in part borrowed from previous work for unsupervised scene decomposition \citep{locatello2020object}, and we use it here to denote hidden representation subspaces of spatiotemporal data.
We aim to bind each slot to a particular component in mixed visual dynamics.
Note that each slot does not directly correspond to a spatiotemporal mode one-to-one.
Instead, slot features can be viewed as latent factors that can explicitly improve the unsupervised decoupling of mixed dynamics across the dataset.
When adapted to a specific sequence, all slots dynamically respond with different importance weights to form compositional representations, which are then used to update the long-term memory state in ModeRNN, termed the \textit{slot bus}.
Specifically, the slot bus is initialized from a learnable, multi-variate Gaussian distribution, whose mean and variance encode the global priors for the entire dataset.

The spatiotemporal slots and the corresponding slot bus are organized in a hierarchical structure, which leads to a better understanding of the complex and highly mixed dynamic patterns without mode annotations, thus providing a way to prevent spatiotemporal mode collapse.
In other words, based on the slot features, the model is allowed to learn similar representation structures from similar data samples. In contrast, it shows significant differences in the learned importance weights as well as the states of slot bus in response to distinct visual dynamics.
The next question is how to extract and separate the slot features from rather chaotic visual sequences. Furthermore, another question is how to build compositional representations based on the slots that can largely differentiate a variety of spatiotemporal modes in a fully unsupervised way.

\subsection{ModeCell}

\begin{figure*}
	\begin{center}
		\centerline{\includegraphics[width=\columnwidth]{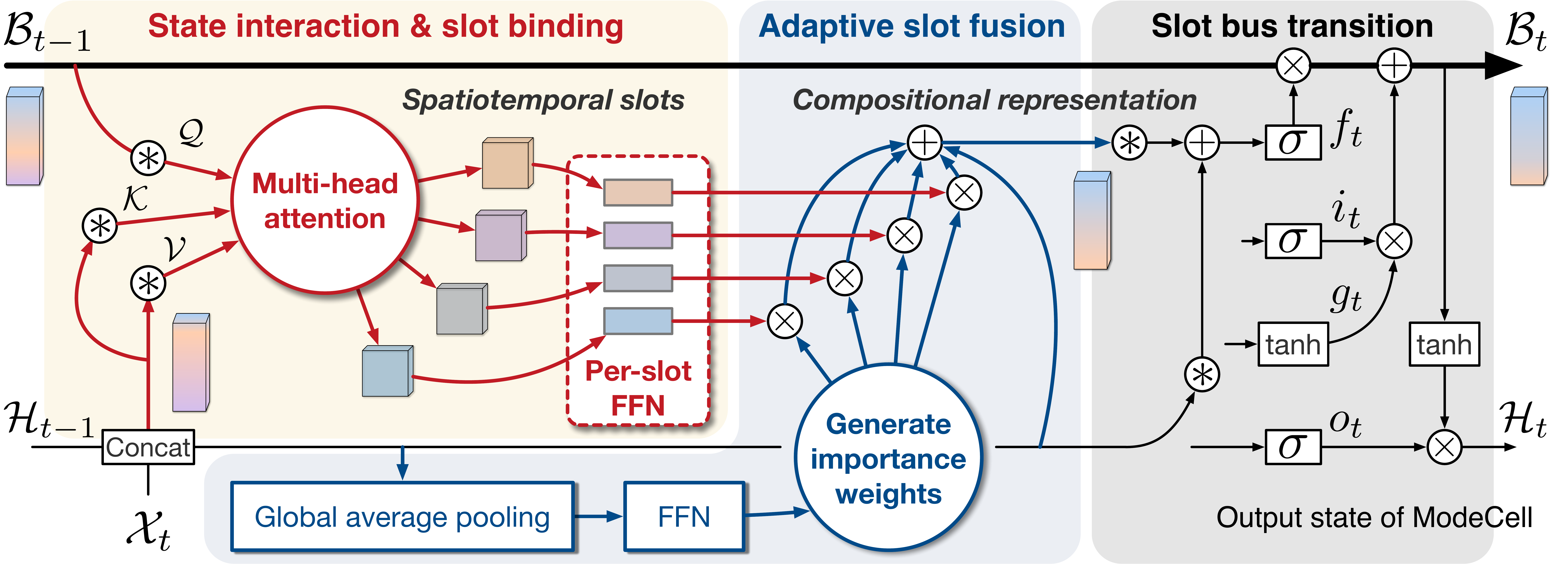}}
		\caption{The internal architecture of ModeCell, which is the basic unit in the proposed ModeRNN.
		}
		\label{fig:ModeCell}
	\end{center}
	\vspace{-10pt}
\end{figure*}

To answer the above questions, we introduce a novel recurrent unit named ModeCell.
It follows a decoupling-aggregation framework (see Figure \ref{fig:ModeCell}) to learn structured hidden representations based on the spatiotemporal slots. Notably, such a framework in the spatiotemporal slot space is completely different from the existing predictive networks with modular architectures \citep{goyal2019recurrent,xu2019unsupervised}.
In ModeCell, the spatiotemporal slots are processed in three stages:
\begin{itemize}
	\item \textbf{State interaction and slot binding:} We first enable the slot bus to interact with the input state and the previous hidden state of ModeRNN through multi-head attention, which naturally improves the decoupling of mixed visual dynamics by dividing output subspaces. To further improves the feature diversity over slots, we then use a feed-forward network (FFN) for each slot to bind the features to independent parameters per slot.
	\item \textbf{Adaptive slot fusion:} We learn to generate importance weights for the decoupled features in each slot, and obtain compositional representations that explicitly consider the co-existence of multiple patterns in a single sequence. The design of this module meets our key argument: on one hand, different modes from the same data domain should share the same set of key factors of spatiotemporal variations; On the other hand, they are encouraged to differ greatly in the combination of these key factors (corresponding to the slots/patterns above).
	\item \textbf{Slot bus transition:} We finally update the slot bus using the compositional representations based on spatiotemporal slots, and then generate the output state of ModeCell. Specifically, we use the LSTM-style gated architectures for the slot bus transitions.
\end{itemize}

Next, we give formalized descriptions of each processing stage in ModeCell.

\myparagraph{State interaction and slot binding.}
Multi-head attention \citep{vaswani2017attention} is widely used in neural language and image processing, and in this work, it is incorporated in the state transitions of ModeRNN.
This mechanism allows interactions between the previous slot bus $\mathcal B_{t-1}$, the current input state $\mathcal X_t$, and the previous hidden state $\mathcal H_{t-1}$ (see Figure \ref{fig:ModeCell}).
Formally, at each time step, we first apply 2D convolution projections to $\mathcal B_{t-1}$. We then flatten the result to 1D and split it into $N$ spatiotemporal slots along the channel dimension, such that $\{\textrm{slot}_{t-1}^{1}, \cdots, \textrm{slot}_{t-1}^{N}\} = \textrm{Split}\left(\textrm{Reshape}\left(W_q \ast \mathcal{B}_{t-1}\right)\right)$.
Note that $\mathcal{B}_{t-1} \in \mathbb{R}^{H \times W \times (d_x+d_h)}$ and $\textrm{slot}_{t-1}^{n} \in \mathbb{R}^{HW(d_x+d_h)/N}$, where $d_x$ is the channel number of input state, $d_h$ is that of hidden state, and $H \times W$ indicates the spatial resolution of the slot bus tensor.
To improve efficiency, we use two $3 \times 3$ depth-wise separable convolutions \citep{chollet2017xception} for $W_q$. 
We use $\{\textrm{slot}_{t-1}^{1}, \cdots, \textrm{slot}_{t-1}^{N}\}$ as the queries of multi-head attention, and apply similar operations to obtain keys $\{\mathcal{K}^{1}_t, \cdots, \mathcal{K}^{N}_t\}$ and values $\{\mathcal{V}^{1}_t, \cdots, \mathcal{V}^{N}_t\}$ based on the concatenation of input state and hidden state, $I_t = [\mathcal{X}_t, \mathcal{H}_{t-1}]$.
We then perform multi-head attention and reshape the $N$ output slot features back to 3D tensors:
\begin{equation}
	\label{equ:MMHA}
	\textrm{slot}_t^{n} = \textrm{Reshape}\left(\textrm{softmax}\left(\frac{\textrm{slot}_{t-1}^{n} \cdot {\mathcal{K}_t^{n}}^{T}}{\sqrt{d_{k}}}\right) \mathcal{V}_t^{n}\right), \ n \in \{1, \cdots, N\}, \\
\end{equation}
where $d_k$ is the dimensionality of the key vectors used as a scaling factor.

Multi-head attention brings two benefits to the forward modeling of spatiotemporal data. 
First, since $\mathcal B_{t-1}$ can be unrolled along the recurrent state transition path to be represented as the transformation of slot features at the previous time step, using $\mathcal B_{t-1}$ as attention queries allows the model to extract features from $\mathcal X_t$ and $\mathcal H_{t-1}$ by jointly attending to prior information at different slots.
Second, the architecture with $N$ attention heads can naturally help factorize the hidden representation into $N$ subspaces, corresponding to $N$ slots.

After the attention module, each slot representation is then independently modeled using per-slot convolutional layers. We update the slot features by
\begin{equation}
	\textrm{slot}_t^{n} = \textrm{FFN}_\textrm{bind}^{n}\left(\textrm{slot}_t^{n}\right), \ n \in \{1, \cdots, N\}. \\
\end{equation}
Through random parameter initialization and stochastic gradient descent, the independent networks $\{\textrm{FFN}_\textrm{bind}^1, \cdots, \textrm{FFN}_\textrm{bind}^N\}$ would most likely be optimized into parameter subspaces far from each other, thus forcing the slots to bind to various components of mixed visual dynamics. 

\myparagraph{Adaptive slot fusion.}

We are motivated by the assumption that different spatiotemporal modes from the same data domain share the same set of slots, and their differences are mainly reflected in the compositional representation based on slot features.
Therefore, we propose to dynamically aggregate $\{\textrm{slot}_{t}^{1}, \cdots, \textrm{slot}_{t}^{N}\}$ with learnable importance weights assigned to each slot.
As shown in Figure \ref{fig:ModeCell}, we first use the global average pooling (GAP) to compress the contextual information in spacetime from $I_t$ to 1D, such that $\textrm{GAP}\left(I_t\right)=\frac{1}{H \times W} \sum_{i=1}^{H} \sum_{j=1}^{W} I_t(i, j)$. 
We then use a feed-forward network with a simple fully-connected layer to reduce the dimensionality and get compact representations. 
Next, we use another group of one-layer fully-connected networks $\{\textrm{FFN}_\textrm{fuse}^1, \cdots, \textrm{FFN}_\textrm{fuse}^N\}$ with independent parameters to generate $N$ sets of importance weights, such that 
\begin{equation}
	\omega_t^{n} = \textrm{FFN}_\textrm{fuse}^{n}\left(\textrm{GAP}\left(I_t\right)\right), \ n \in \{1, \cdots, N\}. \\
\end{equation}
We further use a residual connection from $I_t$ to reduce gradient vanishing, and generate compositional representations $\mathcal F_t$ based on the learned importance weights and corresponding slot features:
\begin{equation}
	\label{equ:slot_fusion}
	\begin{split}
		\mathcal{F}_t & = 
		\textrm{AdaFuse}\left(I_t, \textrm{slot}_t^{1}, \cdots, \textrm{slot}_t^{N}\right) \\ &=\sigma\left(I_t\right)\cdot\left(W_\textrm{fuse}^0 \ast I_t\right) + \sum_{n=1}^{N}\sigma( \omega_t^{n}\cdot I_t)\cdot \left(W_\textrm{fuse}^{n} \ast \textrm{slot}_t^n\right),
	\end{split}
\end{equation}
where $\sigma$ denotes the Sigmoid activation function. $W_\textrm{fuse}^{\ast}$ is a group of $3 \times 3$ convolutional filters.

\myparagraph{Slot bus transition.}

The compositional representation $\mathcal{F}_t$ controls the significance of each slot subspace for a certain sequence at a certain time step.
We use four sets of $\mathcal{F}_t$ to form the input gate $i_t$, forget gate $f_t$, output gate $o_t$, and input modulation gate $g_t$,\footnote{For simplicity, we leave out the gate index in the above discussion. Like standard LSTM, different gates have independent parameters, which applies to all operations before gate generation.} following the gated recurrent transition mechanism from the standard LSTM.
For example, we have $f_t = \sigma\left(W_{ff}\ast\mathcal{F}_t + W_{fi}\ast\textrm{I}_t\right)$
.
Finally, we update the state of slot bus and the output state of ModeCell:
\begin{equation}
	\begin{split}
		&\mathcal{B}_t =f_{t} \odot \mathcal{B}_{t-1}+i_{t} \odot g_{t} \\
		&\mathcal{H}_{t} =o_{t} \odot \tanh (\mathcal{B}_{t}). \\
	\end{split}
\end{equation}

In terms of network architecture, there are multiple ModeCells stacking in ModeRNN. By analogy, the proposed ModeCell is to ModeRNN what LSTM is to the stacked LSTM network.

\section{Experiment}

We use three datasets that are commonly used by previous literature to evaluate ModeRNN. 
\begin{itemize}
	\item \textbf{Mixed Moving MNIST:} It consists of three subsets with $1$--$3$ flying digits, corresponding to three spatiotemporal modes obviously reflected in different frequencies of occlusions. Each subset contains $10{,}000$ training sequences and $3{,}000$ testing sequences. We randomly mix each mode's training sequences, validation sequences, and test sequences to form the overall training, validation, and test sets. Each sequence consists of $20$ consecutive frames, $10$ for input, and $10$ for prediction, in the resolution of $64\times64$.
	\item\textbf{KTH action:} The KTH \citep{Sch2004Recognizing} contains $6$ action categories and involves $25$ subjects in $4$ different scenarios. It thus naturally contains various modes responding to similar action dynamics. We use person $1$-$16$ for training and $17$-$25$ for testing, resize each frame to the resolution of $128\times128$, and predict $20$ frames from $10$ observations.  
	\item \textbf{Radar echo:} This dataset contains $30{,}000$ sequences of radar echo maps for training, and $3{,}769$ for testing. It naturally contains multiple spatiotemporal modes due to seasonal climate (see Appendix \ref{sec:radar_intro}). Models are trained to predict the next $10$ radar echoes based on the previous $10$ observations. All frames are resized to the resolution of $384\times384$.
\end{itemize}
Notably, we do not use any labels in all experiments, because in real-world scenarios, the modes are learned and cannot be pre-defined. In other words, the models are trained in a fully unsupervised way.
We train the models with the $L_{2}$ loss, and use the ADAM optimizer \citep{Kingma2014Adam} with a starting learning rate of $0.0003$.
The batch size is set to $8$, and the training process is stopped after $80{,}000$ iterations. 
%The number of spatiotemporal slots is set to $4$ among all benchmarks. 
All experiments are implemented in PyTorch \citep{PaszkeGMLBCKLGA19} and conducted on NVIDIA TITAN-RTX GPUs. We run all experiments three times and use the average results for quantitative evaluation. 
Typically, we use $4\times 64$-channel stacked recurrent units in most RNN-based compared models, including ConvLSTM, PredRNN, Conv-TT-LSTM, and ModeRNN 
%We provide part of showcases, and more showcases will be provided in the \underline{supplementary material}. 
%

\subsection{Demonstration of Spatiotemporal Mode Collapse}

% \myparagraph{Visualization of spatiotemporal mode collapse.} 

We conduct four experiments to explain spatiotemporal mode collapse: (1) t-SNE visualization of learned features, (2) A-distances of learned features corresponding to different modes, (3) quantitative results of models trained on subsets of pre-defined modes, and (4) showcases of prediction results.

\begin{figure}[t]
	\centering
	\subfigure[t-SNE on Mixed Moving MNIST]{
		\includegraphics[width=0.38\textwidth]{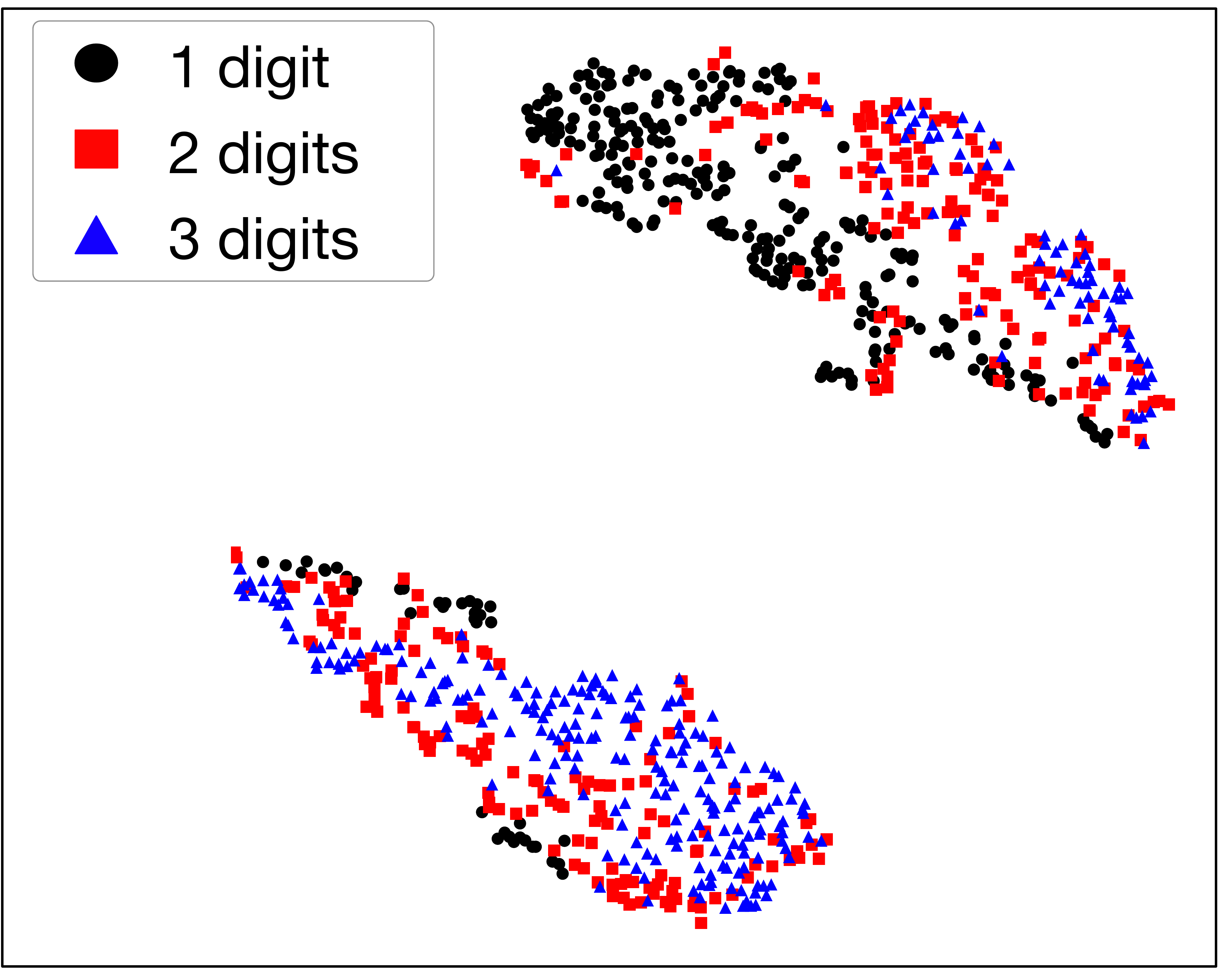}
		\label{fig:tsne_mnist}
	}
	% \vspace{10pt}
	\subfigure[A-distance on KTH]{
		\includegraphics[width=0.58\textwidth]{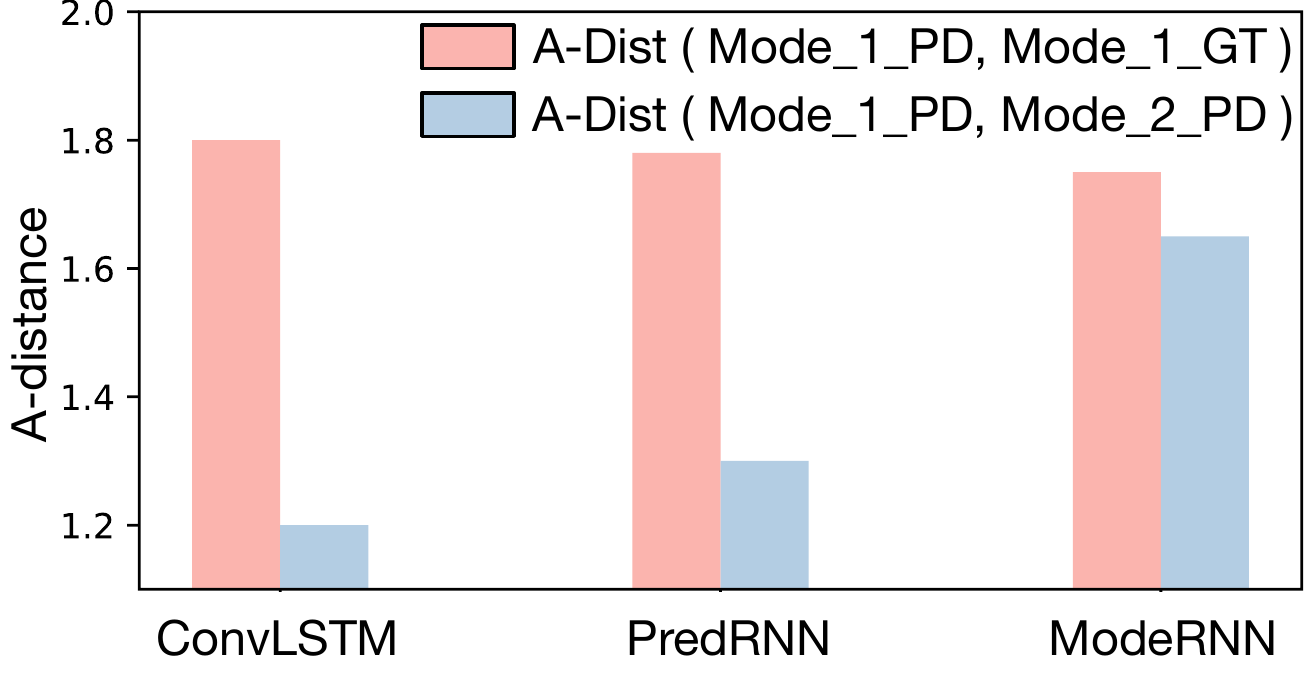}
		\label{fig:a_distance}
	}
	\vskip -0.1in
	\caption{Demonstration of spatiotemporal mode collapse.}
	\vspace{-5pt}
	\label{fig:visual}
\end{figure}

\myparagraph{t-SNE}
In Figure \ref{fig:tsne_mnist}, we visualize the memory state $\mathcal{C}_t$ of ConvLSTM using t-SNE \citep{van2008visualizing}., and find that they are entangled under different digit modes in the Mixed Moving MNIST dataset. 
It provides evidence that this widely used predictive model cannot learn mode structures effectively. 
Training the model on a dataset with mixed dynamics leads to severe mode collapse in feature learning, resulting in the entanglement of hidden representations.

\myparagraph{A-distance.}
A-distance \citep{Ben-DavidBCKPV10} is defined as $d_{A}=2(1-2 \epsilon)$ where $\epsilon$ is the error rate of a domain classifier trained to discriminate two visual domains. 
In Figure \ref{fig:a_distance}, we use the A-distance to quantify the spatiotemporal mode collapse in the real-world KTH action dataset. 
Note that, in this experiment, we divide the KTH dataset into two modes according to the visual similarities of human actions (see details in Appendix \ref{kth:subset}).
As shown by the blue bars (higher is better), the lower A-distance between the two modes indicates that the learned representations from the two modes are highly entangled.
The red bars (lower is better) show the domain distance between features taking as inputs the ground truth frames $\mathcal{X}_t$ and those taking the predictions $\widehat{\mathcal{X}}_t$.
Spatiotemporal mode collapse happens when the A-distance between predictions of different modes (in blue) becomes much smaller than that between predictions and ground truth (in red).
We here use the memory state $\mathcal C_t$ in ConvLSTM and PredRNN, and the slot bus $\mathcal B_t$ in ModeRNN to calculate A-distance.

\myparagraph{Quantitative comparisons of models trained on subset/entire dataset.}
To assess spatiotemporal mode collapse, we separately train predictive models on the subsets of Mixed Moving MNIST (with $1$, $2$, and $3$ digits respectively), and then compare the results with that of the model trained on the mixed dataset. 
As shown in Table \ref{tab:mnist_mode}, previous methods degenerate drastically when using all training samples with mixed visual dynamics.   
The quantitative results perfectly match the visualization in Figure \ref{fig:tsne_mnist}, where the features of Mode-2 and Mode-3 are especially entangled and may influence the learning process of each other.

\begin{figure}[h]
	\centering
	\subfigure[$1$ digit.]{
		\includegraphics[height=0.19\textwidth]{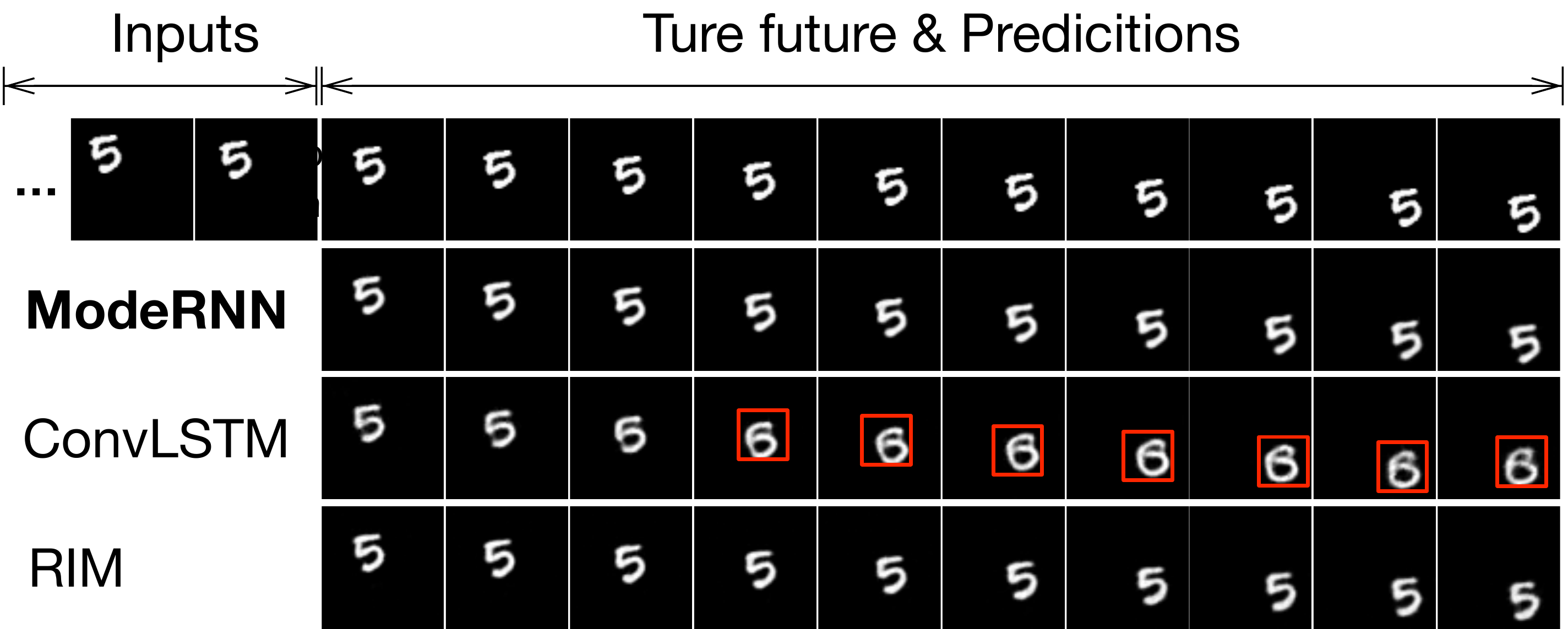}
		% \label{fig:tsne_mnist}
	}
	% \vspace{10pt}
	\subfigure[$3$ digits.]{
		\includegraphics[height=0.19\textwidth]{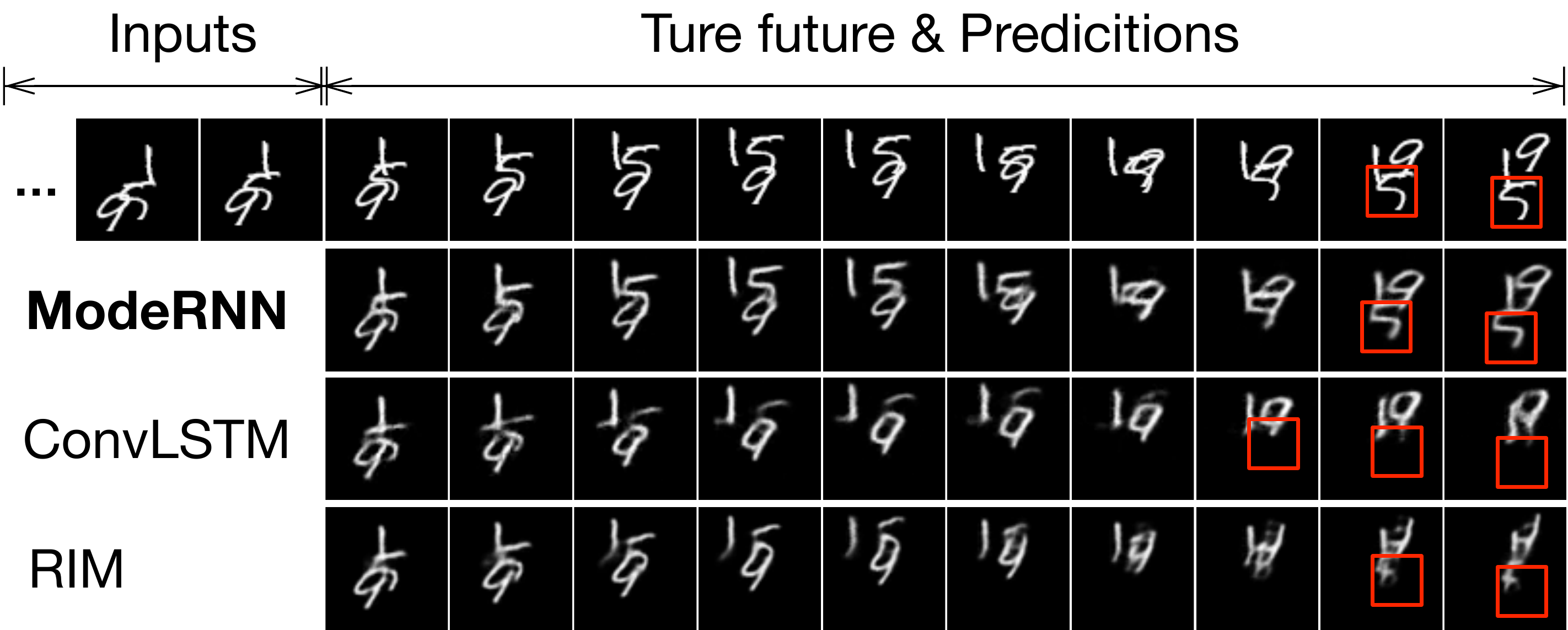}
		%  \label{fig:a_distance}
	}
	\vskip -0.1in
	% \vspace{-10pt}
	\caption{Prediction results on the Mixed Moving MNIST dataset.}
	% \vspace{-15pt}
	\label{fig:mnist_case}
\end{figure}

\myparagraph{Showcases of prediction results.}
As shown in Figure \ref{fig:mnist_case}, we can see that even for the simple case with only $1$ digit, the predicted digit ``5'' from ConvLSTM is gradually twisted across time and even turns into digit ``6'' at last. 
Note that the model is trained on the entire dataset with a variable number of flying digits. The twisted prediction results are caused by the co-existence of the multiple spatiotemporal modes and mode collapse.
In the second showcase in Figure \ref{fig:mnist_case}, the digit ``5'' is even vanished in the predictions of ConvLSTM (indicated by the red box), showing that the 3-digit prediction is collapse to the 2-digit mode.

\begin{table*}[t]
	\caption{Quantitative results of models for each pre-defined mode. We report results by models learned on \textbf{sub/entire} dataset. \textcolor{MyDarkRed}{Red text} indicates improvement by training with entire modes.  \textcolor{MyDarkGreen}{Green text} indicates performance degradation caused by mode collapse. 
	}
	\label{tab:mnist_mode}
	\vskip 0.1in
	\centering
	\begin{scriptsize}
		\begin{sc}
			\renewcommand{\multirowsetup}{\centering}
			\resizebox{1\textwidth}{!}{
				\begin{tabular}{lcccccc}
					\toprule
					\multirow{2}{*}{Model}  & \multicolumn{2}{c}{Mode-1 (1 digit)} & \multicolumn{2}{c}{Mode-2 (2 digits)} & \multicolumn{2}{c}{Mode-3 (3 digits) } \\ 
					\cmidrule(lr){2-7}
					& SSIM ($\uparrow$) & MSE ($\downarrow$) & SSIM ($\uparrow$) & MSE ($\downarrow$) & SSIM ($\uparrow$) & MSE ($\downarrow$)  \\
					\midrule
					ConvLSTM 
					& 0.937 / \textcolor{MyDarkGreen}{0.922}
					& 26.6 / \textcolor{MyDarkGreen}{33.4}
					& 0.848 / \textcolor{MyDarkGreen}{0.839}
					& 63.2 / \textcolor{MyDarkGreen}{65.6}
					& 0.764 / \textcolor{MyDarkGreen}{0.748}
					& 120.5 / \textcolor{MyDarkGreen}{137.1} \\
					%PredRNN (\citet{wang2017predrnn}) 
					%& 0.948 / \textcolor{MyDarkGreen}{0.920} 
					%& 22.8 / \textcolor{MyDarkGreen}{30.1} 
					%& 0.867 / \textcolor{MyDarkGreen}{0.863}
					%& 56.8 / \textcolor{MyDarkGreen}{57.5}   
					%& 0.782 / \textcolor{MyDarkGreen}{0.771}
					%& 93.4 / \textcolor{MyDarkGreen}{114.2}    \\
					%\citet{WangZZLWY19} 
					%& 0.949 / \textcolor{MyDarkGreen}{0.934} 
					%& 24.1 / \textcolor{MyDarkGreen}{28.7} 
					%& 0.872 / \textcolor{MyDarkRed}{0.869}
					%& 57.8 / \textcolor{MyDarkGreen}{59.2}   
					%& 0.802 / \textcolor{MyDarkGreen}{0.773}
					%& 91.4 / \textcolor{MyDarkGreen}{105.3}    \\
					%\citet{su2020convolutional} 
					%& 0.945 / \textcolor{MyDarkGreen}{0.931} 
					%& 22.1 / \textcolor{MyDarkGreen}{27.6} 
					%& 0.874 / \textcolor{MyDarkRed}{0.878}
					%& 52.0 / \textcolor{MyDarkGreen}{55.4}   
					%& 0.817 / \textcolor{MyDarkGreen}{0.788}
					%& 89.2 / \textcolor{MyDarkGreen}{100.6}    \\
					RIM 
					& 0.943 / \textcolor{MyDarkGreen}{0.939} 
					& 23.4 / \textcolor{MyDarkGreen}{25.3} 
					& 0.871 / \textcolor{MyDarkRed}{0.880}
					& 53.3 / \textcolor{MyDarkGreen}{52.1}   
					& 0.814 / \textcolor{MyDarkGreen}{0.803}
					& 88.6 / \textcolor{MyDarkGreen}{95.1}    \\
					\textbf{ModeRNN}
					& 0.946 / \textcolor{MyDarkRed}{\textbf{0.951}} 
					& \textbf{21.9} / \textcolor{MyDarkRed}{\textbf{17.1}} 
					& \textbf{0.880} / \textcolor{MyDarkRed}{\textbf{0.902}} 
					& \textbf{51.0} / \textcolor{MyDarkRed}{\textbf{42.1}} 
					&\textbf{ 0.821} / \textcolor{MyDarkRed}{\textbf{0.842}} 
					& \textbf{83.1} / \textcolor{MyDarkRed}{\textbf{74.8}} \\
					\bottomrule
				\end{tabular}
			}
		\end{sc}
	\end{scriptsize}
	% \end{small}
	% \vspace{-5pt}
\end{table*}

\subsection{Visualization of Representations Learned by ModeRNN} 

\begin{figure}[t]
	\centering
	\subfigure[Spatiotemporal slots]{
		\includegraphics[height=0.26\textwidth]{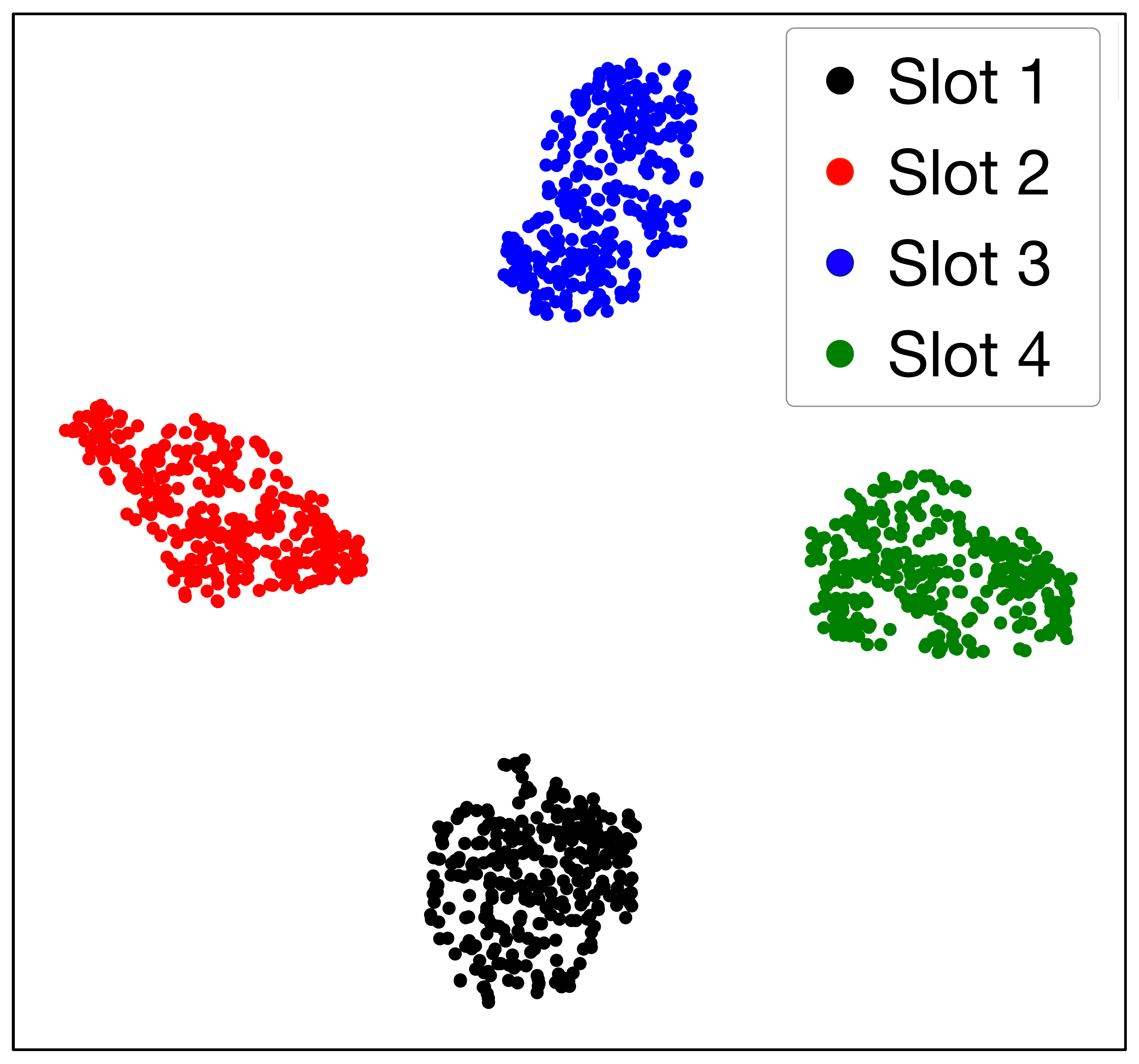}
		\label{fig:visual_slots}
	}
	\subfigure[Slot bus]{
		\includegraphics[height=0.26\textwidth]{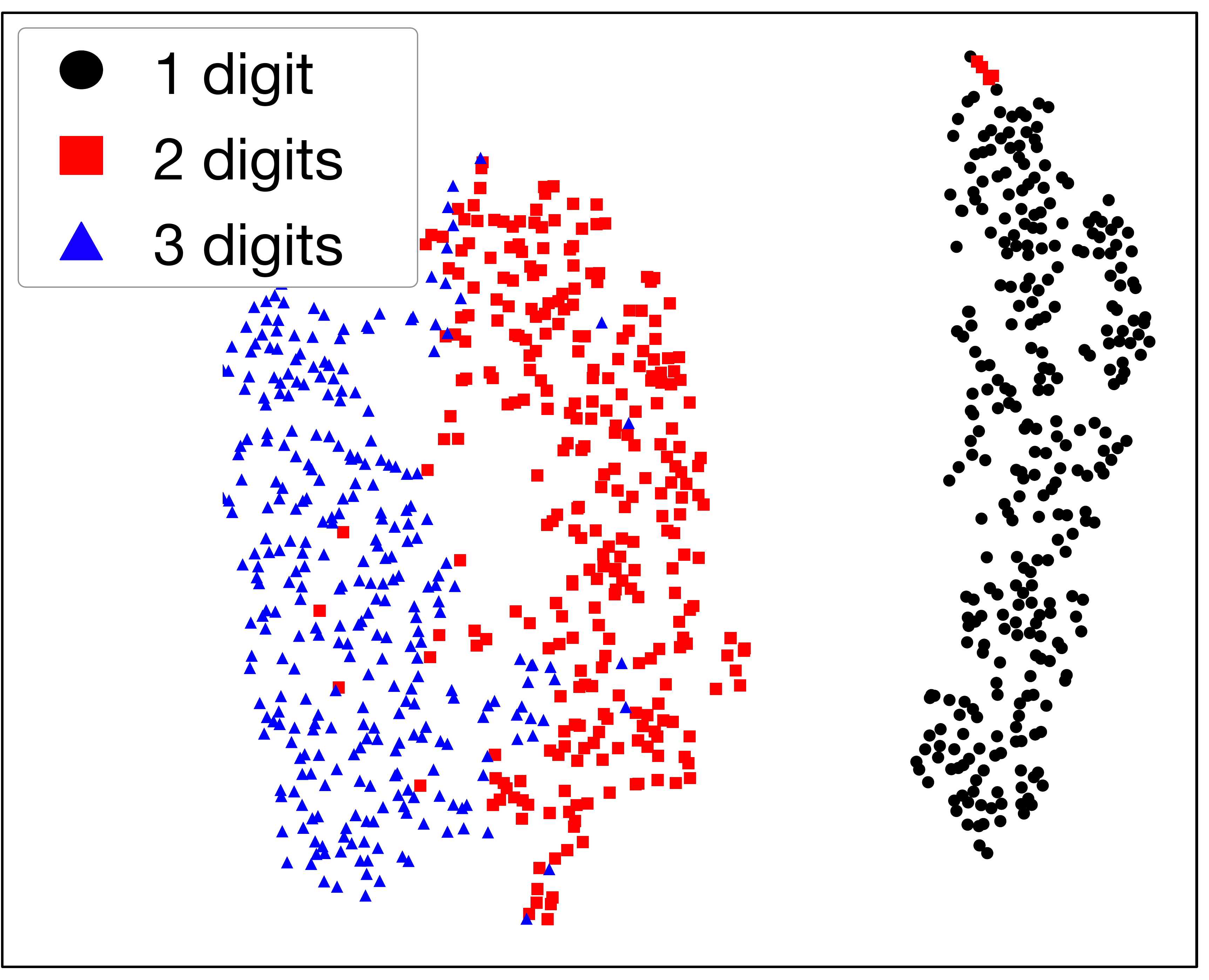}
		\label{fig:visual_slot_bus}
	}
	\subfigure[Importance weights]{
		\includegraphics[height=0.26\textwidth]{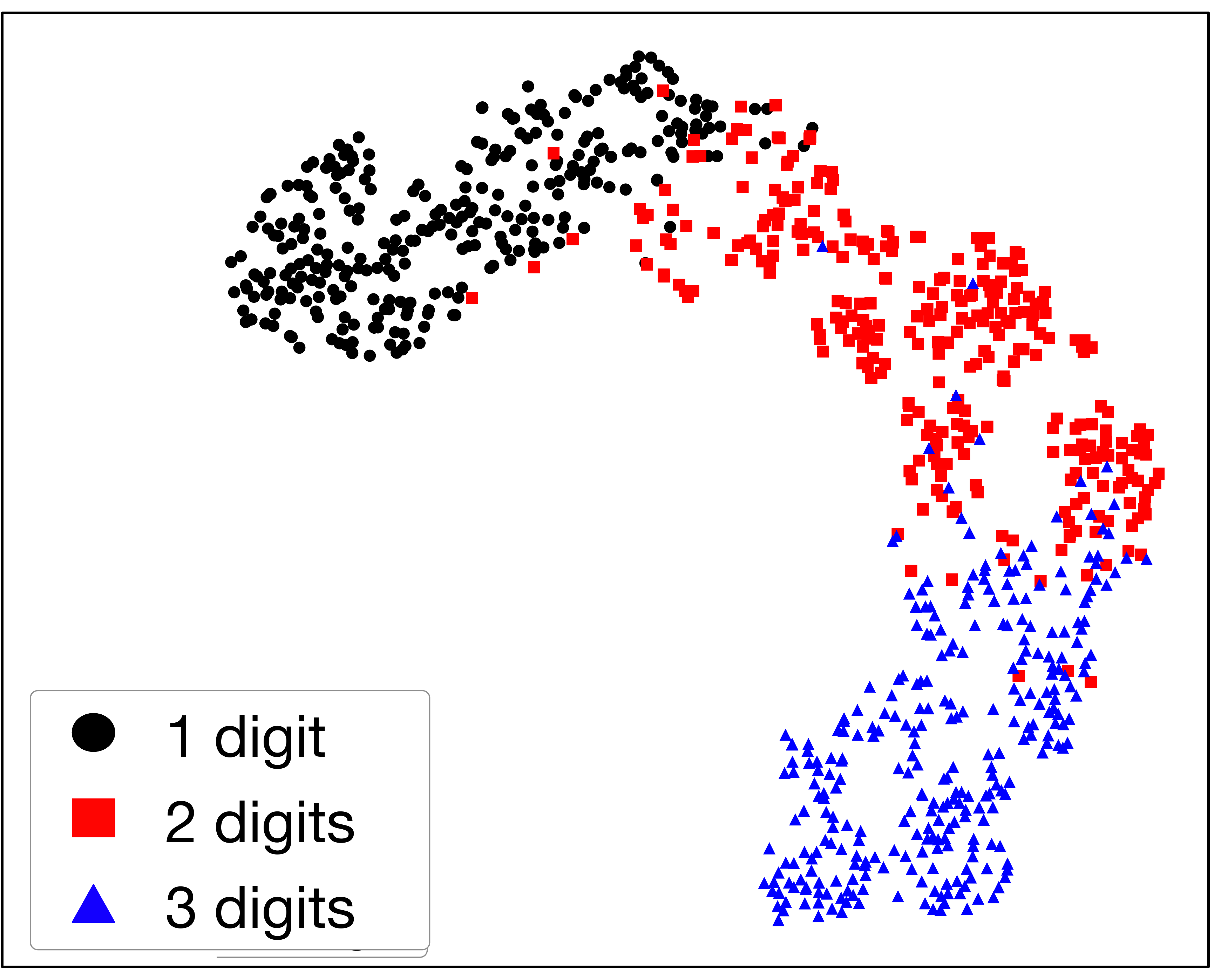}
		\label{fig:visual_important_weigths}
	}
	\vskip -0.1in
	\caption{Visualization of representations learned by ModeRNN on the Mixed Moving MNIST.}
	\label{fig:visual}
	% \vspace{-10pt}
\end{figure}

To testify the mode decoupling ability of our ModeRNN, we visualize the slot features in Figure \ref{fig:visual_slots}. Here, we use $4$ spatiotemporal slots.
We can see that these slot features are obviously clustered into exact $4$ groups, indicating that different slots of ModeRNN learn diverse dynamic patterns from highly mixed spatiotemporal modes. 
In Figure \ref{fig:visual_slot_bus}, we further visualize the features in the slot bus, which show $3$ clusters with clear boundaries, corresponding to the three modes with different numbers of digits. 
In Figure \ref{fig:visual_important_weigths}, we visualize the importance weights of samples in different modes.
The above results indicate that, first, the spatiotemporal slots are successfully bound to different components of visual dynamics; Second, the compositional representations based on the slots, including the slot but, successfully learn discriminative knowledge from different dynamic modes.

\subsection{Resuls of ModeRNN on Mixed Moving MNIST}

\myparagraph{Main results.} 
Table \ref{tab:mnist_mode} gives the results on each subset of flying digit mode. We can see that only ModeRNN achieves consistently better performance on each mode. It is also the only one that can consistently benefit from training with mixed dynamics from the entire dataset. 
Besides, in Table \ref{tab:mnist}, we show the overall quantitative results as well as computational efficiency of the compared models on the Mixed Moving MNIST dataset. 
As we can see, ModeRNN achieves \textbf{state-of-the-art} overall performance (SSIM: \textbf{0.897}, MSE: \textbf{44.7}) with \textbf{fewer parameters} compared with existing approaches, including the more recent work \citep{goyal2019recurrent} also with modular architectures.
Furthermore, as shown in Figure \ref{fig:mnist_case}, ModeRNN is the only method that can capture the exact movement of each digit, while other models predict the blurry results and the digit ``5'' is even vanished. All in all, our ModeRNN could effectively overcome the spatiotemporal mode collapse to achieve competitive performance on complex datasets.

\myparagraph{Hyperparameters analysis on the number of slots.} In Table \ref{tab:ablation2}, we gradually increase the number of spatiotemporal slots from $1$, finding that the performance of ModeRNN increases rapidly at the early stage and achieves the best performance on $N=4$. We have similar results on all benchmarks and set $N=4$ throughout this paper. 
%We provide further ablation study on the adaptive slot fusion module and the slot binding module in Appendix \ref{sec:ab}.

%\myparagraph{Ablation Study on Model Components}
%\label{sec:ab}

%We analyze the efficacy of each part in ModeRNN on the mixed Moving MNIST dataset, as shown in Table \ref{tab:ablation3}. Without our adaptive mixture, the model could not work well (MSE: $44.7 \rightarrow 60.5$). It strongly proves that the adaptive mixture is crucial for combining the latent dynamic modes to overcome the spatiotemporal mode collapse.
%We also evaluate the signal decomposition effect, which cooperates with filter bank to capture latent dynamic patterns. Using the original input without decomposition could not effectively cope with mixing dynamics (MNIST-3 MSE: $82.3 \rightarrow 117.3$). Thus, both the signal decomposition cooperating with the filter bank and the adaptive mixture are indispensable to ModeRNN.

\myparagraph{Ablation study on model components.} We analyze the efficacy of each network component in ModeRNN on the Mixed Moving MNIST dataset, as shown in Table 4. Without the adaptive slot fusion module, the model can not work well by simply adding different slot features with equal weights (MSE: 44.7 → 60.5), which strongly demonstrates that learning adaptive, compositional representations over the slot features is crucial to modeling the underlying structure of visual dynamics and can therefore better decouple the learned spatiotemporal modes.

\begin{table*}[t]
	\vspace{-10pt}
	\caption{Quantitative results and parameter analysis on the Mixed Moving MNIST dataset. Here, we report the training time for $1{,}000$ sequences. 
	}
	\label{tab:mnist}
	\vskip 0.1in
	\centering
	\begin{small}
		\begin{sc}
			\renewcommand{\multirowsetup}{\centering}
			\resizebox{\textwidth}{!}{
				\begin{tabular}{lcccccccc}
					\toprule
					\multirow{1}{*}{Model}  
					& SSIM ($\uparrow$) & MSE ($\downarrow$) & Param (MB) & Time (s) & Mem (GB) \\
					\midrule
					ConvLSTM \citep{shi2015convolutional} 
					& 0.836
					& 78.7  & 8.15 & 29 & 2.58\\
					PredRNN \citep{wang2017predrnn}
					& 0.851 
					& 67.3  & 11.83 & 68 & 3.52\\
					MIM \citep{WangZZLWY19} 
					& 0.851
					& 64.4  & 18.45 & 73 & 4.22\\
					Conv-TT-LSTM \citep{su2020convolutional} 
					& 0.866
					& 61.2 & 7.58  & 77 & 6.17\\
					RIM \citep{goyal2019recurrent}
					& 0.874
					& 57.5  & 7.91  & 71 & 5.88\\
					\textbf{ModeRNN}
					& \textbf{0.898}
					& \textbf{44.7} & \textbf{6.35} & 59 & 3.15\\
					\bottomrule
				\end{tabular}
			}
		\end{sc}
	\end{small}
	% \vspace{-10pt}
\end{table*}

\begin{table}[h]
	\vspace{-10pt}
	\caption{The ablation study of ModeRNN with respect to the number of spatiotemporal slots. 
	}
	\label{tab:ablation2}
	\centering
	\small
	\vskip 0.05in
	\begin{sc}
		\renewcommand{\multirowsetup}{\centering} 
		\setlength{\tabcolsep}{3.7pt}
		\scalebox{1.0}{
			\begin{tabular}{lcccccccc}
				\toprule
				\multirow{1}{*}{Model}  & \multicolumn{1}{c}{Per-frame MSE} \\
				\midrule
				ModeRNN ($N=1$) & 69.3    \\ 
				ModeRNN ($N=2$) & 59.4  \\
				ModeRNN ($N=3$) & 48.2 \\
				\textbf{ModeRNN ($N=4$, final proposed)} &\textbf{44.7}  \\
				ModeRNN ($N=5$) &45.9   \\
				\bottomrule
			\end{tabular}
		}
	\end{sc}
	% \vspace{-10pt}
\end{table}

\begin{table}[h]
	\caption{The ablation study on the adaptive slot fusion module and the slot binding module on the mixed Moving MNIST dataset.}
	\label{tab:ablation3}
	\centering
	\vskip 0.15in
	\begin{sc}
		\renewcommand{\multirowsetup}{\centering} 
		\setlength{\tabcolsep}{3.7pt}
		\scalebox{1.0}{
			\begin{tabular}{lcccccccc}
				\toprule
				\multirow{1}{*}{Model}  & \multicolumn{1}{c}{Per-frame MSE} \\
				\midrule
				\textbf{ModeRNN} &\textbf{44.7}   \\
				ModeRNN w/o Slot Binding ($N=1$) & 69.3    \\
				ModeRNN w/o Adaptive Slot Fusion & 60.5  \\
				\bottomrule
			\end{tabular}
		}
	\end{sc}
\end{table}

\subsection{Resuls of ModeRNN on the KTH Action Dataset}

%\myparagraph{Visualization of spatiotemporal mode collapse and representations learned by ModeRNN.}

% \myparagraph{Main results.}

%According to the scale of the actions, we can simply group the existing six categories in the KTH dataset into two typical modes:
%\begin{itemize}[leftmargin=*]
%    \item The first mode corresponds to the global movement of the torso, including the categories of running, walking, and jogging.
%    \item The second mode corresponds to the local movement of hands, including the categories of hand-clapping, hand waving, and boxing.
%\end{itemize}

\begin{figure}[h]
	\begin{center}
		\centerline{\includegraphics[width=0.9\columnwidth]{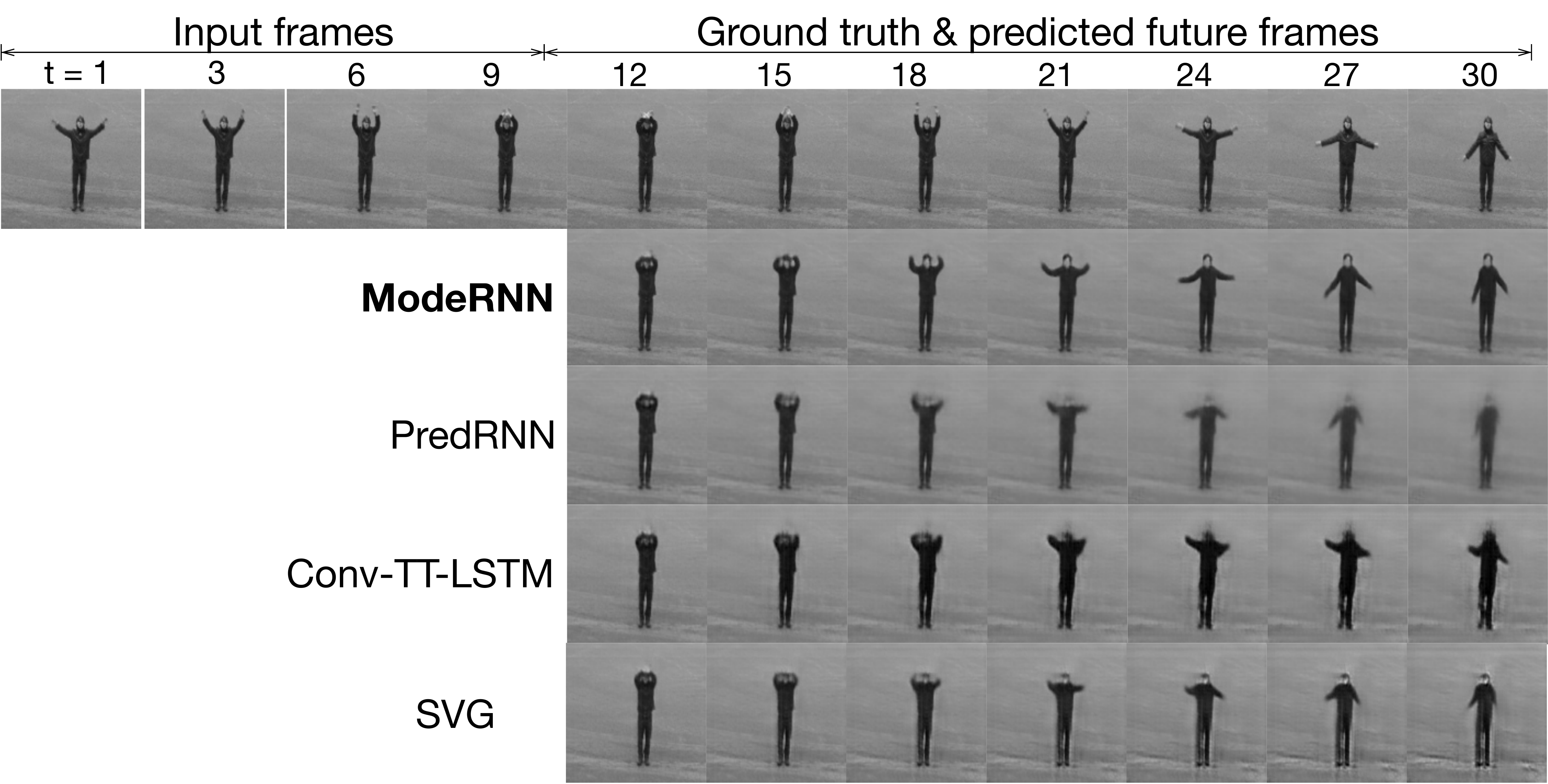}}
		\vspace{-5pt}
		\caption{Prediction frames on the KTH action dataset.}
		\label{fig:kth_case}
	\end{center}
	%\vspace{-10pt}
\end{figure}

On this dataset, we use the frame-wise peak signal-to-noise ratio (PSNR) and learned perceptual image patch similarity (LPIPS) \citep{zhang2018unreasonable} as evaluation metrics.
In the left column of Table \ref{tab:kth_and_guangzhou}, we show the quantitative results of the KTH dataset and find that ModeRNN achieves better overall performance among all compared methods. Impressively, ModeRNN overcomes the mode collapse effectively and achieves better performance compared with the powerful probabilistic model SVG \citep{denton2018stochastic} (PSNR : 27.73 vs \textbf{28.22}, LPIPS : 0.196 vs \textbf{0.183}).  We provide the qualitative showcases in Figure  \ref{fig:kth_case} where we can see that ModeRNN can predict the precise position of the moving person.

\begin{table*}[t]
	% \vskip -0.1in
	\caption{Quantitative results on the KTH action dataset and the radar echo dataset. For the probabilistic SVG model, we report the best results from $100$ output samples per input sequence.
	}
	\label{tab:kth_and_guangzhou}
	\vskip 0.1in
	\centering
	\begin{small}
		\begin{sc}
			\renewcommand{\multirowsetup}{\centering}
			\begin{tabular}{lcccc}
				\toprule
				\multirow{2}{*}{Model} & \multicolumn{2}{c}{KTH} & \multicolumn{2}{c}{Radar}\\
				\cmidrule(lr){2-5}
				& PSNR ($\uparrow$) & LPIPS ($\downarrow$) & CSI30 ($\uparrow$) & MSE ($\downarrow$)\\
				\midrule
				ConvLSTM \citep{shi2015convolutional} &  24.12 & 0.231 & 0.354 & 97.6  
				\\
				TrajGRU \citep{shi2017deep} & 26.97 & 0.219 & 0.357 & 89.2 \\
				PredRNN \citep{wang2017predrnn}  & 27.47  & 0.212 & 0.359 & 84.2  \\
				SVG \citep{denton2018stochastic} 
				&  27.73 & 0.196 & - & - \\
				Conv-TT-LSTM \citep{su2020convolutional} &  27.59 &  0.198  & 0.363 & 87.6 \\
				\textbf{ModeRNN}  & \textbf{28.22} & \textbf{0.183} & \textbf{0.428} & \textbf{65.1} \\
				\bottomrule
			\end{tabular}
		\end{sc}
	\end{small}
	%\vspace{-10pt}
\end{table*}

\subsection{Resuls of ModeRNN on the Radar Echo Dataset}
\label{sec:radar}

Besides the frame-wise MSE, we use the Critical Success Index (CSI) metric, which is defined as $\textrm{CSI}=\frac{\textrm{Hits}}{\textrm{Hits}+\textrm{Misses}+\textrm{FalseAlarms}}$, where hits correspond to the true positive, misses correspond to the false positive, and false alarms correspond to the false negative. A higher CSI indicates better forecasting performance, and it is particularly sensitive to high-intensity echoes. We set the alarm threshold to $30$ dBZ for this radar benchmark. 
As shown in the right column of Table \ref{tab:kth_and_guangzhou}, ModeRNN achieves the \textbf{state-of-the-art} overall performance and significantly outperforms the competitive precipitation method, TrajGRU \citep{shi2017deep} (CSI: \textbf{0.428} vs 0.357, MSE: \textbf{65.1} vs 89.2).

\begin{figure}[h]
	% \vspace{-5pt}
	\begin{center}
		\centerline{\includegraphics[width=1\columnwidth]{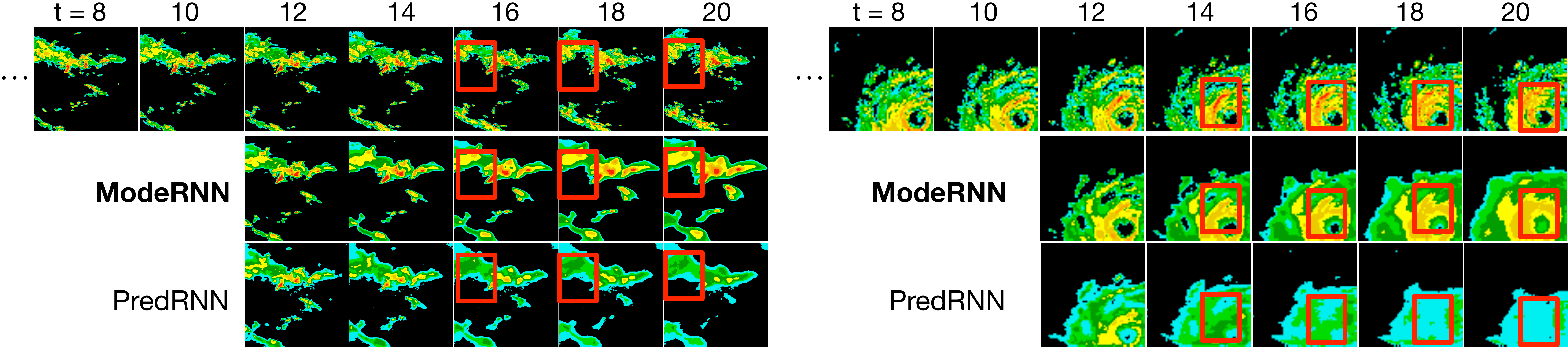}}
		% 		\vspace{-5pt}
		\caption{Prediction results on the radar echo dataset.}
		\label{fig:guangzhou_case}
	\end{center}
	\vspace{-10pt}
\end{figure}

As shown in Figure \ref{fig:guangzhou_case}, we find that the compared models fail in predicting the edges of the cyclone, and the predicted movement of the cloud even vanishes. On the contrary, ModeRNN provides more details for the cyclone and accurately predicts its center position indicated by the red box.
Both the quantitative and qualitative results show that our ModeRNN can effectively capture the dynamic information from complex meteorological dynamic modes.

\section{Conclusion}

In this paper, we demonstrated a new experimental phenomenon of spatiotemporal mode collapse when training unsupervised predictive models on a dataset with highly mixed visual dynamics. 
Accordingly, we proposed a novel predictive network named ModeRNN, which can effectively learn modular features from the mixed dataset using a set of spatiotemporal slots.
To discover the compositional structures in spatiotemporal modes, ModeRNN adaptively aggregates the slot features with learnable importance weight.
Compared with existing models, ModeRNN was shown to prevent the mode collapse of future predictions in spacetime, improving qualitative and quantitative results on three widely used datasets.

\bibliography{iclr2022_conference}
\bibliographystyle{iclr2022_conference}

\clearpage
\appendix

\section{Illustration of Mixed Modes for Precipitation Forecasting}
\label{sec:radar_intro}

\begin{figure}[h]
	\centering
	\includegraphics[width=\textwidth]{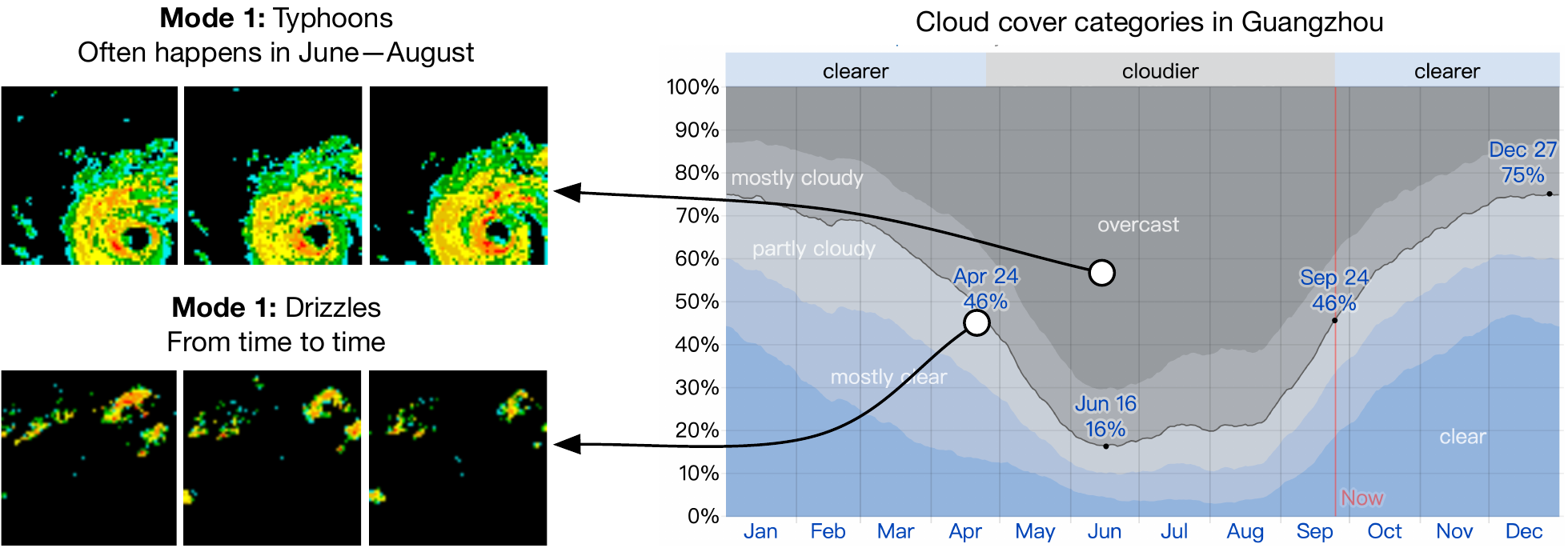}
	\vskip -0.05in
	\caption{Illustration of mixed spatiotemporal modes in the radar echo dataset. In practice, we cannot pre-assign mode labels before forecast because of the fuzzy boundaries between seasonal climates.
	}
	\label{fig:weather}
\end{figure}

Real-world spatiotemporal datasets usually contain a variety of latent modes of visual dynamics without human annotations.
As discussed in Section \ref{sec:radar}, we take precipitation forecasting based on radar echo observations as a practical application scenario of spatiotemporal prediction.
Figure \ref{fig:weather} gives an example of the mixed dynamics for the typical climate in Guangzhou, China. 
Obviously, since the climate changes smoothly between seasons with fuzzy boundaries, it is inappropriate to pre-assign mode labels before forecast. 

\section{Additional Results and Discussion}

% \subsection{Ablation Study on the Number of Spatiotemporal Slots}

\subsection{Training Models on Subsets Divided by Pre-Defined Mode Labels}

\paragraph{Subsets in KTH based on pre-defined action categories.}
\label{kth:subset}

According to the scale of the actions, we can simply group the existing six categories in the KTH dataset into two typical modes:
\begin{itemize}[leftmargin=*]
	\item The first mode corresponds to the global movement of the torso, including the categories of running, walking, and jogging.
	\item The second mode corresponds to the local movement of hands, including the categories of hand clapping, hand waving, and boxing.
\end{itemize}

\begin{table*}[h]
	\caption{Quantitative results of models that are learned on the \textbf{sub/entire} KTH action dataset. For the probabilistic SVG model \citep{denton2018stochastic}, we report the best results from $100$ output samples per input sequence. 
	}
	\label{tab:kth}
	\vskip 0.1in
	\centering
	\begin{scriptsize}
		\begin{sc}
			\renewcommand{\multirowsetup}{\centering}
			\resizebox{1\textwidth}{!}{
				\begin{tabular}{lcccccc}
					\toprule
					\multirow{2}{*}{Model}  & \multicolumn{2}{c}{Mode-1 (torso movement)} & \multicolumn{2}{c}{Mode-2 (hand movement)} & \multicolumn{2}{c}{Overall}\\ 
					\cmidrule(lr){2-7}
					& PSNR ($\uparrow$) & LPIPS ($\downarrow$) & PSNR ($\uparrow$) & LPIPS ($\downarrow$) & PSNR ($\uparrow$) & LPIPS ($\downarrow$) \\
					\midrule
					\citet{shi2015convolutional} & 23.65 / \textcolor{MyDarkGreen}{23.13} & 0.227 / \textcolor{MyDarkGreen}{0.232} & 24.37 / \textcolor{MyDarkRed}{24.98} & 0.234 / \textcolor{MyDarkRed}{0.230} &  24.03 / \textcolor{MyDarkRed}{24.12} & 0.231 / 0.231  
					\\
					\citet{wang2017predrnn} & 27.45 / \textcolor{MyDarkGreen}{26.92} & 0.218 / \textcolor{MyDarkGreen}{0.226} & 27.52 / \textcolor{MyDarkRed}{27.94} & 0.203 / \textcolor{MyDarkRed}{0.199}  &27.49 / \textcolor{MyDarkGreen}{27.47}  & 0.210 / \textcolor{MyDarkGreen}{0.212}  \\
					\citet{denton2018stochastic} 
					& 27.79 / \textcolor{MyDarkGreen}{27.34} &   0.193 / \textcolor{MyDarkGreen}{0.199}   & 27.75 / \textcolor{MyDarkRed}{28.12} & 0.197 / \textcolor{MyDarkRed}{0.193} & 27.77 /  \textcolor{MyDarkGreen}{27.73} & 0.195 / \textcolor{MyDarkGreen}{0.196} \\
					\citet{su2020convolutional} & 27.57 / \textcolor{MyDarkGreen}{27.09} & 0.192 / \textcolor{MyDarkGreen}{0.204} & 27.64 / \textcolor{MyDarkRed}{28.03} & 0.201 / \textcolor{MyDarkRed}{0.192} & 27.61 / \textcolor{MyDarkGreen}{27.59} &  0.196 / \textcolor{MyDarkGreen}{0.198}  \\
					%TMU \cite{yao2020unsupervised} & -&- &- &- & 0.777 & 94.7\\
					\textbf{ModeRNN}  & 27.84 / \textcolor{MyDarkRed}{\textbf{28.11}} & 0.192 / \textcolor{MyDarkRed}{\textbf{0.185}} & 27.82 / \textcolor{MyDarkRed}{\textbf{28.32}} & 0.196 / \textcolor{MyDarkRed}{\textbf{0.181}} & 27.83 / \textcolor{MyDarkRed}{\textbf{28.22}} &  0.194 / \textcolor{MyDarkRed}{\textbf{0.183}} \\
					\bottomrule
				\end{tabular}
			}
		\end{sc}
	\end{scriptsize}
	% \end{small}
\end{table*}

\paragraph{Subsets in the radar echo dataset based on pre-defined chronological climate modes.}

Considering the climate change between different seasons in Guangzhou, we can roughly divide the radar echo dataset into two typical meteorology groups:
\begin{itemize}[leftmargin=*]
	\item The first mode: It corresponds to the windier part of the year, from March to May, with average wind speeds of more than $7.5$ miles per hour. There will be drizzles from time to time in these months. We use the radar maps from $2016/3$ to $2016/5$ and from $2017/3$ to $2017/4$ for training, and use those in $2017/5$ for testing. 
	\item The second mode: It corresponds to the the summer in Guangzhou, which from Figure \ref{fig:intro}, experiences heavier cloud cover, with the percentage of time that the sky is overcast or mostly cloudy is around $80\%$.
	We use the radar maps from $2016/6$ to $2016/8$ and from $2017/6$ to $2017/7$ for training, and use those in $2017/8$ for testing. 
\end{itemize}

\begin{table*}[h]
	\caption{Quantitative results of models that  are learned on the \textbf{sub/entire} radar echo dataset. \textcolor{MyDarkRed}{Red text} indicates improvement by joint training.  \textcolor{MyDarkGreen}{Green text} indicates performance degradation caused by mode collapse. 
	}
	\vskip 0.1in
	\centering
	\begin{scriptsize}
		\begin{sc}
			\renewcommand{\multirowsetup}{\centering}
			\resizebox{1\textwidth}{!}{
				\begin{tabular}{lcccccc}
					\toprule
					\multirow{2}{*}{Model}  & \multicolumn{2}{c}{Mode-1 (Mar.--May)} & \multicolumn{2}{c}{Mode-2 (Jun.--Aug.)} & \multicolumn{2}{c}{Overall}\\ 
					\cmidrule(lr){2-7}
					& CSI30 ($\uparrow$) & MSE ($\downarrow$) & CSI30 ($\uparrow$) & MSE ($\downarrow$) & CSI30 ($\uparrow$) & MSE ($\downarrow$) \\
					\midrule
					\citet{shi2015convolutional} & 0.341 / \textcolor{MyDarkGreen}{0.337} & 63.5 / \textcolor{MyDarkGreen}{71.3} & 0.372 / \textcolor{MyDarkGreen}{0.366} & 102.4 / \textcolor{MyDarkGreen}{116.7}   
					&  0.359 / \textcolor{MyDarkGreen}{0.354} 
					& 86.0 / \textcolor{MyDarkGreen}{97.6} 
					\\
					\citet{shi2017deep} & 0.354 / \textcolor{MyDarkGreen}{0.341} & 62.5 / \textcolor{MyDarkGreen}{68.3} & 0.375 / \textcolor{MyDarkGreen}{0.369} & 99.3 / \textcolor{MyDarkGreen}{104.3} 
					&  0.366 / \textcolor{MyDarkGreen}{0.357} 
					& 83.8 / \textcolor{MyDarkGreen}{89.2} \\
					\citet{wang2017predrnn} & 0.357 / \textcolor{MyDarkGreen}{0.342} & 57.1 / \textcolor{MyDarkGreen}{60.8} & 0.384 / \textcolor{MyDarkGreen}{0.371} & 97.5 / \textcolor{MyDarkGreen}{101.2}   
					&  0.373 / \textcolor{MyDarkGreen}{0.359} 
					& 80.5 / \textcolor{MyDarkGreen}{84.2} \\
					\citet{su2020convolutional}& 0.359 / \textcolor{MyDarkGreen}{0.347} & 61.5 / \textcolor{MyDarkGreen}{65.8} & 0.381 / \textcolor{MyDarkGreen}{0.374} & 99.1 / \textcolor{MyDarkGreen}{103.5}   
					&  0.372 / \textcolor{MyDarkGreen}{0.363} 
					& 83.3 / \textcolor{MyDarkGreen}{87.6} \\
					\textbf{ModeRNN} & 0.373 / \textcolor{MyDarkRed}{\textbf{0.408}} & 54.3 / \textcolor{MyDarkRed}{\textbf{46.2}} & 0.392 / \textcolor{MyDarkRed}{\textbf{0.442}} & 92.0 / \textcolor{MyDarkRed}{\textbf{78.8}} 
					&  \textbf{0.384} / \textcolor{MyDarkRed}{\textbf{0.428}} 
					& \textbf{76.1} / \textcolor{MyDarkRed}{\textbf{65.1}} \\
					\bottomrule
				\end{tabular}
			}
		\end{sc}
	\end{scriptsize}
	\label{tab:radar}
\end{table*}

\end{document}